\documentclass[conference]{IEEEtran}  

\IEEEoverridecommandlockouts                              

\usepackage[utf8]{inputenc}
\usepackage[T1]{fontenc}
\usepackage{graphicx}
\graphicspath{ {./figures} }
\usepackage[nospace,noadjust]{cite}
\usepackage{textcomp}
\usepackage[T1]{fontenc}
\usepackage[english]{babel}
\usepackage[section]{placeins}

\usepackage{amsfonts}
\usepackage{color}
\usepackage[cmex10]{amsmath}

\title{\LARGE \bf
Activity Monitoring of Islamic Prayer (Salat) Postures using Deep Learning
}

\author{\IEEEauthorblockN{Anis Koubaa\IEEEauthorrefmark{1},\IEEEauthorrefmark{5},\IEEEauthorrefmark{3},
Adel Ammar\IEEEauthorrefmark{1}, Bilel Benjdira\IEEEauthorrefmark{1}, \IEEEauthorrefmark{6}, Abdullatif Al-Hadid\IEEEauthorrefmark{1}, Belal Kawaf\IEEEauthorrefmark{1}, \\Saleh Ali Al-Yahri\IEEEauthorrefmark{1}, Abdelrahman Babiker, Koutaiba Assaf, Mohannad Ba Ras }\\

\IEEEauthorblockA{\IEEEauthorrefmark{1}Robotics and Internet-of-Things Lab (RIOTU), Prince Sultan University, Riyadh, Saudi Arabia.\\
}
\IEEEauthorblockA{\IEEEauthorrefmark{2} CISTER Research Centre, ISEP, Polytechnic Institute of Porto, Porto, Portugal\\}
\IEEEauthorblockA{\IEEEauthorrefmark{5}Gaitech Robotics, China. }
\IEEEauthorblockA{\IEEEauthorrefmark{6}Research Laboratory SEICT, LR18ES44. National Engineering School of Carthage,  Tunisia.\\ } \\
Email: (akoubaa, aammar, bbenjdira)@psu.edu.sa

}

\begin{document}

\maketitle
\thispagestyle{empty}
\pagestyle{empty}

\begin{abstract}
In the Muslim community, the prayer (\textit{i.e.} Salat) is the second pillar of Islam, and it is the most essential and fundamental worshiping activity that believers have to perform five times a day. From a gestures' perspective, there are predefined human postures that must be performed in a precise manner. However, for several people, these postures are not correctly performed, due to being new to Salat or even having learned prayers in an incorrect manner. Furthermore, the time spent in each posture has to be balanced. To address these issues, we propose to develop an artificial intelligence assistive framework that guides worshippers to evaluate the correctness of the postures of their prayers. This paper represents the first step to achieve this objective and addresses the problem of the recognition of the basic gestures of Islamic prayer using Convolutional Neural Networks (CNN). The contribution of this paper lies in building a dataset for the basic Salat positions, and train a YOLOv3 neural network for the recognition of the gestures. Experimental results demonstrate that the mean average precision attains 85\% for a training dataset of 764 images of the different postures. To the best of our knowledge, this is the first work that addresses human activity recognition of Salat using deep learning. 

\end{abstract}


\section{INTRODUCTION}

Human Activity Recognition has been extensively investigated using different techniques including sensing technologies \cite{alobaid2018prayer, al2016prayer, Rabbi2019}, computer vision \cite{Ramzan2019}, and more recently using deep learning \cite{Ravi2017, Xu2019}. It consists of the classification of the activity of a person from the data collected from sensors (\textit{e.g.} accelerometer, camera, and laser scanner ). The advances in human activity recognition have enabled different applications in different areas such as healthcare, sports activities, violence detection, older people monitoring, postures' recognition, to name a few. 

In this paper, we consider a particular human activity application with special interest to the Muslim community around the world, namely the recognition of postures of the Islamic prayer, also known as \textit{Salat}. Salat is the second pillar in Islam and is the most important worshipping activity that is repeated five times a day by any Muslim. Besides its spiritual value, it consists of a series of postures that must be executed in a predefined sequence as instructed by the Prophet Mohamed, peace be upon him. The motivation of this work is that several people (e.g., kids, beginners) are likely not to perform the postures correctly. There are different reasons for this to happen like, for example, the lack of proper learning of prayer movements, or being careless. Besides, another requirement in Salat is the \textit{tranquility}; that is, every posture should be executed in a proper and sufficient amount of time while making invocations and reading the Quran. Only a few research works in the literature have attempted to address the activity recognition of Salat. In \cite{alobaid2018prayer, al2016prayer}, the authors used smartphone technology to recognize Salat activities. In \cite{Rabbi2019, Ibrahim2012}, the authors assessed the activities of Salat using Electromyographic (EMG) signals.
On the other hand, several research works used deep learning approaches to address the problem of human activity recognition in different applications such as sports, healthcare, well-being using wearable sensors \cite{Ravi2017, Xu2019, Gumaei2019}. To the best of our knowledge, there is no previous work that addressed the problem of activity monitoring of Islamic prayer using deep learning approaches. In this paper, we propose a novel solution to the posture recognition of Salat using convolutional neural networks (CNN), which aims at identifying the four basic postures of Salat namely, Standing (Qiam), Bowing (Ruku), Prostration (Sujud), Sitting (Julus), using state-of-the-art CNN algorithms. CNN has been used in a wide variety of applications such as vehicle detection \cite{benjdira2019conf, Ammar2019}, semantic segmentation of urban environments \cite{Benjdira2019Segmentation}, and self-driving vehicles \cite{schoettle2014survey}.
The contribution presented in this paper represents the first step towards the main objective of this project that consists in developing an AI-based tool for the assessment of the Islamic prayer and an assistive system that helps beginners and kids to correct the postures during Salat. It has to be noted that we only consider the recognition of correct postures in our neural network model, and we ignore wrong positions, which is left as a future extension of this work in which we will address the anomalies during prayers. The principal added value of this paper is the development of a specific dataset for the human activities in the context of Islamic prayer that provides four classes related to each of the basic postures of Salat. Furthermore, we trained the state-of-the-art YOLOv3 neural network on the constructed dataset to accurately detect and recognize the four postures of Salat. We have tested the resulting trained network on videos showing people praying, and all the postures were most of the time correctly recognized. 
The remainder of the paper is organized as follows. Section II discusses the literature on human activity recognition and also related works on Islamic prayer monitoring using sensing activities. Section III gives a brief overview of the YOLOv3 algorithm. Section IV presents the different postures of Salat and the corresponding dataset. Section V discusses the experimental results. Finally, Section VI concludes the paper and presents future extensions.

\section{RELATED WORKS}
The earliest work that targets the task of automatic recognition of prayer movements is the paper introduced in 2009 by El-Hoseiny et al.\cite{ElHoseiny2009MuslimPA}. They used a camera to capture the side view of the prayer. Then, they used morphological operations to extract the polygon corresponding to the contour of the prayer body. The polygon information is used to calculate the backbone axis angle and to extract four key-points of the human body. The backbone axis angle is the angle between the y-axis and the back-axis of the detected body. The four key points determined from the polygon are the center point of the polygon, the ankle point, the head point, and the back point. Based on the coordinates of the key-points and the value of the backbone axis angle, the prayer postures and movements can be determined using a set of given inequations. The method is totally based on hand-crafted features without any use of machine learning classifiers. This work is the only work that used ordinary camera sensor for this task. The other works leveraged the use of accelerometer and Kinect sensors. Al Ghannam et al.\cite{al2016prayer} used a set of classifiers for prayer activity monitoring and recognition based on accelerometer data collected from a mobile phone. They used three machine learning classifiers:   J48 Decision Trees, IB1 (Instance-Based Learning ) Algorithm, and Naive Bayes. The accuracy of all these algorithms exceeded 90\%. Eskaf et al.\cite{eskaf2016aggregated} presented a framework for daily life activities (sitting, standing, walking, ...) from accelerometer data. Then, they aggregated these simple activities for recognition of prayer using supervised machine learning classifiers. Ali et al.\cite{ali2018salat} presented similarly a system for automatic recognition and monitoring of prayer postures based on triaxial accelerometer data collected from a smartphone. They added an analysis of group prayer activities using dynamic time warping. 

In \cite{Ravi2016}, the authors proposed a human activity recognition technique based on a deep learning model designed for low power devices. The recognition gives significant contextual data for well-being. Alobaid et al.\cite{alobaid2018prayer} also studied the use of mobile accelerometer data for prayer activity recognition. They made a performance comparison of three feature extraction approaches and eight machine learning classifiers. They concluded that Random Forest is the most appropriate for this task, with an accuracy of 90\%. They proposed a two-level classifier to solve the confusion between two similar prayer stages and to improve the accuracy to 93\%. They added a brief study on the effect of personal characteristics like height and age on the performance of the classifier. Jaafer et al.\cite{jaafar2015investigation} used another sensor, the Kinect RGB-Depth camera. Two Kinect sensors are put in a fixed position on the body, and the skeleton information is extracted using Kinect Software Development Kit. They used the Hidden Markov Model as the machine learning classifier for learning the prayer movements from the skeleton information. \par
We can see from the above works that only one work had used the ordinary camera sensor, although its practicability and easiness of use compared to other sensors. Moreover, this work used only geometric properties without profiting from the potential of machine learning algorithms. The significant advance noted in machine learning and especially the emergence of deep learning algorithms will surely make a significant improvement in the efficiency of prayer postures recognition from a video camera. This is why we targeted this limitation in the current state of the art approaches. We selected one of the most performing one-stage deep learning algorithms in object detection task (YOLOv3\cite{YOLOv3}), and we applied it in the context of prayer posture recognition. We considered in our study that the camera could be placed in different views from the prayer and not only in a fixed position from him.

\section{ALGORITHMS BACKGROUND}
Among the deep learning algorithms used in computer vision, YOLOv3\cite{YOLOv3} is the most attractive for the task of prayer posture recognition. Two facts are defending this choice. Firstly, YOLOv3\cite{YOLOv3} had proven its efficiency compared to other object detection algorithms\cite{benjdira2019conf}. Secondly, it has an efficient inference time (up to 45 frames per second). This allows a real-time recognition of the prayer postures. In the following subsections, we present the architecture of the first version of YOLO (YOLOv1\cite{YOLO2016}) and the different improvements implemented in YOLOv2\cite{YOLOv2} and YOLOv3\cite{YOLOv3}.

\subsection{YOLOv1}
\begin{figure*}  
\begin{center}
\includegraphics[width=18cm]{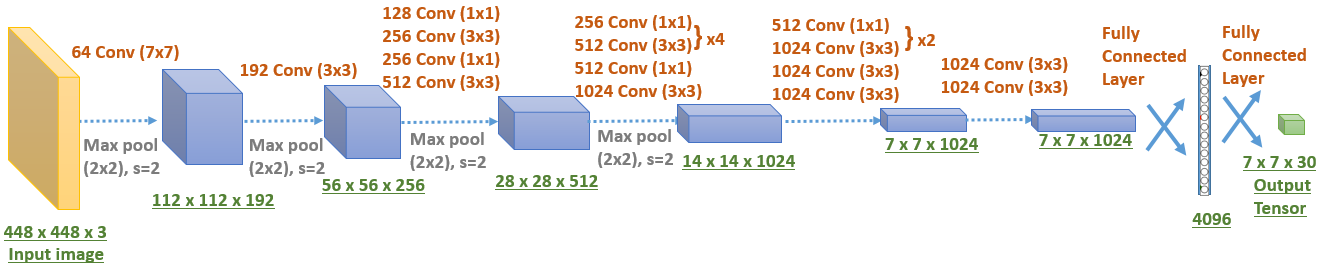}
\caption{\small \sl YOLO v1 Architecture} \label{fig:yolo-architecture}  
\end{center}  
\end{figure*}

YOLO was first introduced in 2016 as a different approach to treat generic object detection problems. The approach is based on training only one CNN for the task of localization and the task of classification at the same time. 
YOLOv1 is composed of 24 convolutional layers for feature extraction and 2 fully connected layers for generating the final output. The total architecture is detailed in \(Figure\) \ref{fig:yolo-architecture}. 
 \par
The input image is divided by YOLO into a grid of \(S\times S\). Every grid cell can be associated with only one object. The grid cell has also a fixed number \(B\) of boundary boxes to be predicted for this object. A confidence score is calculated for every bounding box. As a result, YOLO calculates for every grid cell a vector of class probabilities among the \(C\) classes we are targeting. YOLO calculates also for every bounding box of the cell a vector containing 5 parameters: \((x,y,w,h,box\_confidence\_score)\). Hence, the YOLO network generates for every input image a tensor of the form:
\begin{equation} \label{eq:1}
     S\times S\times (B*5+C)  
\end{equation}
where:
\begin{itemize}
\item \( S\times S\): corresponds to the number of grid cells.
\item \( B\): corresponds to the number of bounding boxes.
\item \(C\): corresponds to the number of targeted classes.
\end{itemize}
To train the YOLO network, we used a combination of three loss functions. The first is the classification loss (loss of the conditional probabilities for every class). The second is the localization loss (the position of the estimated bounding box compared to the ground truth). The third is the confidence loss (the box confidence score compared to the ground truth). To measure the error between the predicted value and the ground truth, YOLO uses sum-squared error as a metric. The expression  of the loss is detailed in \(Equation\) \ref{eq:2}. 

\begin{equation} \label{eq:2}
\begin{gathered}
Loss = \lambda_{coord}\sum_{i=0}^{S^2} \sum_{j=0}^{B} {\rm 1\!l}_{ij}^{obj}[(x_i-\hat{x}_i)^2 +(y_i-\hat{y}_i)^2] \\
+ \lambda_{coord}\sum_{i=0}^{S^2} \sum_{j=0}^{B} {\rm 1\!l}_{ij}^{obj}[(\sqrt{\omega_i}-\sqrt{\hat{\omega}_i})^2 +(\sqrt{h_i}-\sqrt{\hat{h}_i})^2] \\
+ \sum_{i=0}^{S^2} \sum_{j=0}^{B} {\rm 1\!l}_{ij}^{obj}(C_i- \hat{C}_i)^2 \\
+ \lambda_{noobj}\sum_{i=0}^{S^2} \sum_{j=0}^{B} {\rm 1\!l}_{ij}^{noobj}(C_i- \hat{C}_i)^2 \\
+ \sum_{i=0}^{S^2}{\rm 1\!l}_{i}^{obj} \sum_{c \in classes}(p_i(c) -\hat{p}_i(c))^2
\end{gathered}
\end{equation}
where:
\begin{itemize}
   \item \(\lambda_{coord} \) is the weight for the bounding box coordinate prediction loss. During the training, it is set to 5.
    \item \(\lambda_{noobj} \) is the weight for the confidence score prediction loss for boxes that don't include objects. During the training, It is set to 0.5.
    \item \({\rm 1\!l}_{i}^{obj}\) indicates if the object appears in cell \(i\)
    \item  \({\rm 1\!l}_{i}^{obj}\) indicates that the bounding box of index \(j\) is the responsible for the prediction.
    \item \(x_{i}\) is the ground truth value of \(x\), \(\hat{x}_i\) is the predicted value for \(x\), The same for \(y\), (\ \(\omega\) (width) and \(h\) (height). 
    \item \(C\) corresponds to the confidence score of the bounding box
    \item \(p_{i}(c)\) is the probability that the cell \(i\) belong to the class \(c\). \(\hat{p}\) denotes the predicted probability. 
    
\end{itemize}
When it was introduced, YOLO surpassed other object detection algorithms in terms of speed. Its mAP (mean Average Precision) was comparable or exceeded the mAP of other state-of-the-art algorithms. 
\subsection{YOLOv2}
In YOLOv2\cite{YOLOv2}, many improvements had been introduced to increase the accuracy and decrease the processing time. Among them, we can note:
\begin{itemize}
    \item Use of Batch Normalization (BN). This technique was introduced in 2015 to improve the convergence of the loss during the training. BN was added to all convolutional layers in YOLO, which improved the mAP by 2\%. 
    \item Replacing the input image size \(224 \times 224 \) by \(448 \times 448\), which improved the mAP by 4\%. 
    \item The adoption of convolution with anchor boxes. The class prediction is moved from the level of the grid cell to the level of the boundary box. This made a small increase in the mAP by a margin of 0.3\% but increased the recall from 81\% to 88\%. This enhanced the ability to detect all the objects existing in the image and reduced the False Negative Rate.
    \item The use of K-means clustering algorithm to select the best anchor box from the training set of bounding boxes. Euclidian distance is replaced by the IoU (Intersection Over Union) for the clustering.
    \item The predictions are made on the offsets to the anchors. Instead of predicting \((x, y, \omega, h, C)\), YOLOv2 predicts \((t\textsubscript{x},t\textsubscript{y},t\textsubscript{w},t\textsubscript{h},t\textsubscript{C})\). This makes the network converge better.  
    \item The use of fine-grained features. Similarly to the identity mapping in ResNet \cite{ResNet}, YOLOv2 concatenates low resolution features to high resolution features to improve the ability to detect small objects. This improves the mAP by 1\%. 
    \item The adoption of multi-scale training. Instead of using a fixed size for the input image, YOLOv2 randomly selects an image size every 10 batches. This improves the ability to predict well over different sizes of input images.  
\end{itemize}
\subsection{YOLOv3}
In April 2018, the YOLOv3\cite{YOLOv3} was introduced as an incremental improvement to the previous versions. Among the improvements made, we can note:
\begin{itemize}
    \item The use of the multi-label classification. Instead of the mutual exclusive labeling in the previous versions, YOLOv3 uses a logistic classifier to estimate the likeliness of the object being of a specific label. The classification loss is changed to use for every label the binary cross-entropy loss instead of the general mean square loss used in the previous versions. 
    \item The use of another bounding box prediction. During the training, the objectness score 1 in YOLOv3 is associated with the bounding box anchor that best overlaps the ground truth object. Besides, if the IoU (Intersection Over Union) between the bounding box anchor and the ground truth is less than a threshold (0.7 in the implementation), it is ignored. In the end, YOLOV3 associates for every ground truth object one bounding box anchor. 
    \item The change of the output 3d tensor. In YOLOv3, the prediction is done for one grid cell at 3 different scales and then concludes the final bounding box from those scales. This was inspired by the feature pyramid networks. Hence, the new dimension of the output 3d sensor is then:
    \begin{equation} \label{eq:1}
     S\times S\times (3*(5+C))  
    \end{equation}
    where:
    \begin{itemize}
    \item \( S\times S\): corresponds to the number of grid cells.
    \item \(B\): is omitted because only one bounding box anchor is kept at the end. 
    \item \(C\): corresponds to the number of targeted classes.
    \end{itemize}
    \item The adoption of a new feature extractor (the darknet-53). It has 53 layers and uses skip connection similarly to ResNet \cite{ResNet}. It uses both \(3 \times 3\) and \(1 \times 1\) convolutions. It gave the state of the art accuracy but with better speed and fewer computations. 
\end{itemize}

\section{DATASET}

\begin{figure}
\begin{center}
\includegraphics[width=\linewidth, height=8cm]{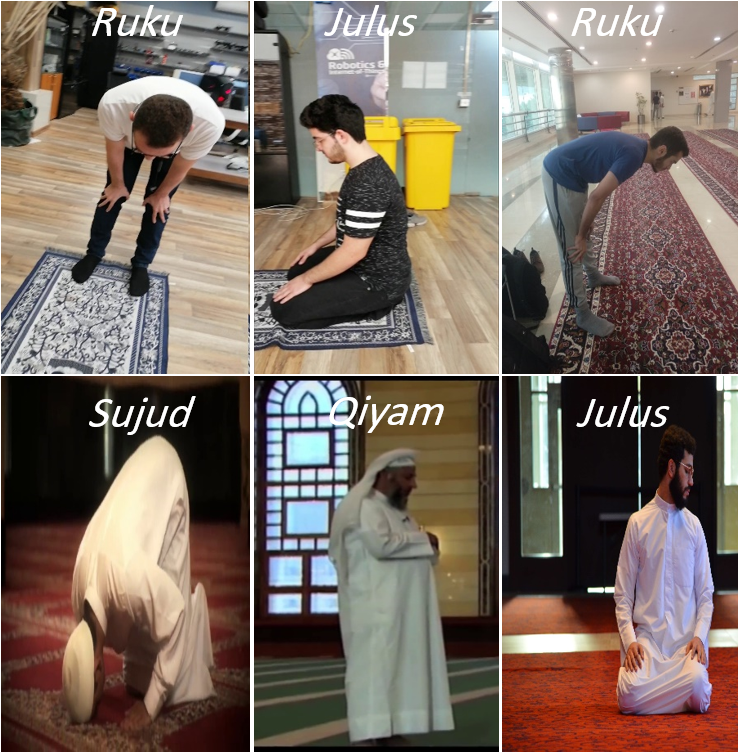}
\caption{Sample images of our dataset, showing the different prayer positions. The images on the top were captured specifically for this study, while the images on the bottom were collected from the Internet.
\label{fig:dataset-sample}}  
\end{center}  
\end{figure}

We have built the Salat Postures Dataset from a set of images of students and laboratory members performing different Salat positions using mobile phones, from various view angles, especially captured for the purpose, complemented with images (and video frames) of people in prayer collected from the Internet. In fact, our ultimate scenario is that a person records himself using a mobile phone, and then the video stream will be processed in real-time to identify the different postures and finally provides postures flaws and recommendation. For this reason, most of the collected images are based on mobile phone data collection to serve the final scenario, added to this some more general images. We have also considered images that show the whole body of the person making prayer, as we impose that the person records the whole body from any angle to perform the detection and classification. 

The total number of collected images is 849 that we manually labeled into four classes using rectangular bounding boxes. The four classes correspond to the four main positions in Islamic prayer: 

\begin{itemize}
    \item \textit{Qiyam}: it is the act of standing, and this posture is related to when reading the Quran, or after rising from bowing (\textit{Ruku}). It typically spans over one or more minutes. Sub-figure 5 of Figure 2 shows an example of a man in \textit{Qiyam} posture. 
    \item \textit{Ruku}: it means bowing, and this posture happens after Qiam to make invocations. It typically spans over a few seconds. Sub-figures 1 and 3 of Figure 2 shows an example of a man in \textit{Ruku} posture.  
    \item \textit{Sujud}: It is the prostration and happens after raising from bowing for also making invocation. It spans over several seconds. Sub-figure 4 of Figure 2 shows an example of a man in \textit{Sujud} posture.  
    \item\textit{Julus}: it is the position of sitting after finishing \textit{Sujud} for making invocation and closing Salat at its end. Sub-figures 2 and 6 of Figure 2 shows an example of a man in \textit{Julus} posture.     
\end{itemize}

Figure \ref{fig:dataset-sample} shows some sample images of the dataset, and Table \ref{tab:Dataset} shows the number of images and instances in the training and testing datasets. We have subdivided the training data and test data to be 90\% and 10\%, respectively.

\begin{table}[]
\caption{Number of images and instances in the training and testing datasets.}
\label{tab:Dataset}
\begin{tabular}{lcc}
\hline
                                 & \multicolumn{1}{l}{Training dataset} & \multicolumn{1}{l}{Testing dataset} \\ \hline
Number of images                 & 764                                  & 85                                  \\ \hline
Percentage                       & 90.0\%                               & 10.0\%                              \\ \hline
Instances of class "Standing"    & 303                                  & 37                                  \\ \hline
Instances of class "Bowing"      & 210                                  & 27                                  \\ \hline
Instances of class "Prostrating" & 174                                  & 11                                  \\ \hline
Instances of class "Sitting"     & 179                                  & 22                                  \\ \hline
\end{tabular}
\end{table}

\section{EXPERIMENTAL EVALUATION}

\begin{figure}[h]
\begin{center}
\includegraphics[width=\linewidth]{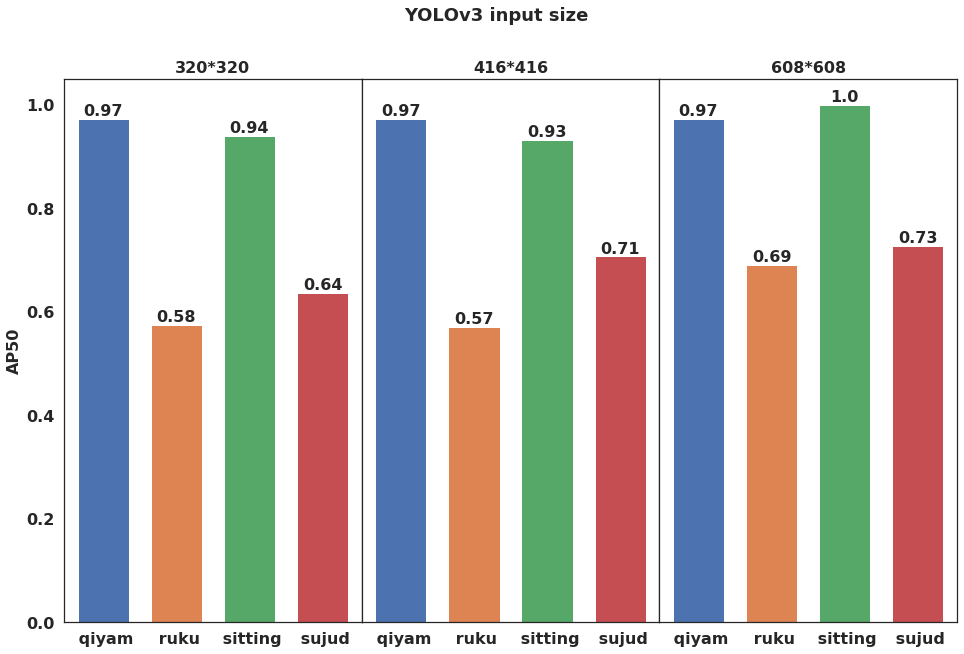}
\caption{Average Precision (AP) at an IoU threshold of 0.5, for each input size and each class.\label{fig:AP-per-class}}  
\end{center}  
\end{figure}

For the experimental setup, we used a workstation powered by an Intel Core  i9-9900K
(Octa-core) processor, with 64GB RAM, and an NVIDIA GeForce RTx2080T
(11 GB) GPU, running on Ubuntu 16.04 LTS.

We used the following metrics to assess the performance of the object detection networks:

\begin{itemize}
    \item IoU: Intersection over Union, which measures the overlap between the predicted bounding box and the ground-truth one.
    \item mAP: mean average precision, which corresponds to the area under the precision vs. recall curve, averaged over all classes. The mAP was measured for different values of IoU (0.5, 0.6, 0.7, 0.8 and 0.9).
    \item Inference time (in millisecond per image), which measures the inference processing speed on test images.
    \item TP (True Positives): number of objects correctly detected and classified.
    \item FP (False Positives): number of objects detected but wrongly classified.
    \item FN (False Negatives): number of non-detected objects.
\end{itemize}

\begin{figure}[h]
\begin{center}
\includegraphics[width=\linewidth]{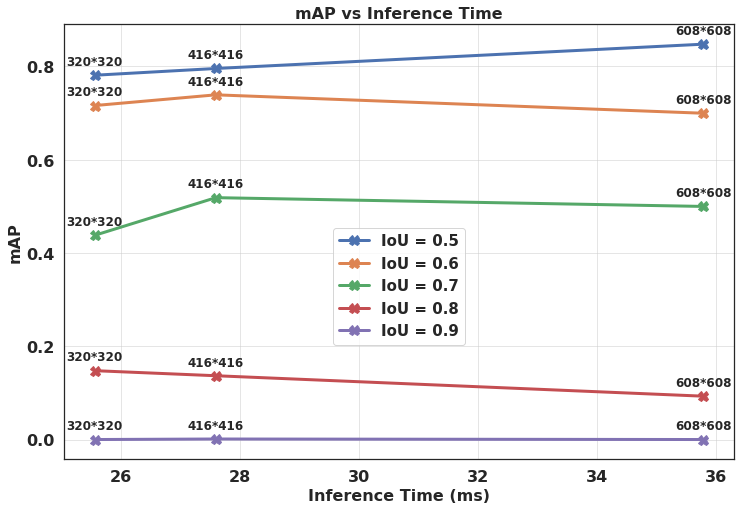}
\caption{Comparison of the trade-off between mAP and inference time for the three different input sizes of YOLO v3, and for different values of IoU threshold ranging from 0.5 to 0.9.\label{fig:AP-Inference-time}}  
\end{center}  
\end{figure}

\begin{figure}[h]
\begin{center}
\includegraphics[width=7cm]{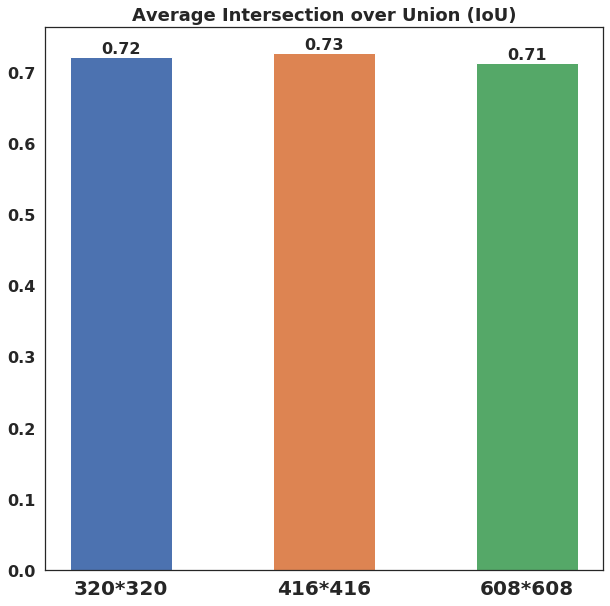}
\caption{Average intersection over Union (IoU) for the three input sizes. \label{fig:IoU}}  
\end{center}  
\end{figure}

\begin{figure}[h]
\begin{center}
\includegraphics[width=\linewidth]{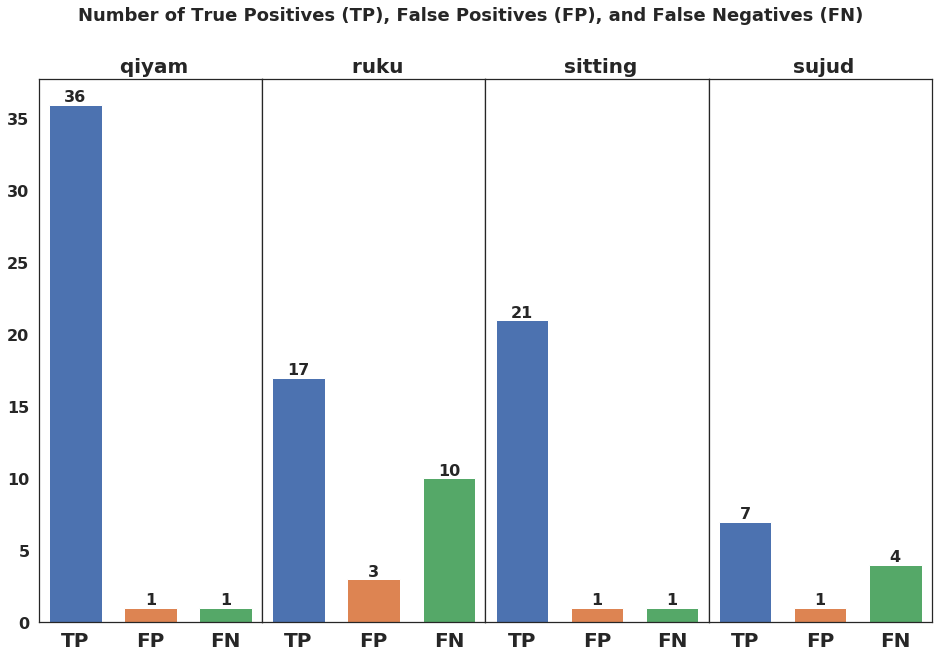}
\caption{Number of true positives (TP), false positives (FP), and false negatives (FN) for an IoU threshold of 0.5, and an input size of 320x320. \label{fig:TP-FP-FN}}  
\end{center}  
\end{figure}

Figure \ref{fig:AP-per-class} shows the value of AP50 (Average Precision for an IoU threshold of 0.5), for each input size and each class. For all three networks, the performance obtained for the detection of \textit{Ruku} and \textit{Sujud} classes are markedly lower than for qiyam and sitting classes. This may be explained by the visual resemblance between these two classes, which causes them to be conflated.  

Figure \ref{fig:AP-Inference-time} displays the trade-off between mAP and inference time for the 3  input sizes, and for different values of IoU threshold ranging from 0.5 to 0.9. For an IoU threshold of 0.5, passing from an input size of 320*320 to 416*416, the mAP50 increases by 2\%, while the inference time increases by 8\%. Then, when passing from an input size of 416*416 to 608*608, the mAP50 increases by 7\%, while the inference time jumps by 30\%. For higher values of IoU threshold, increasing the input size has little or negative effect on the mAP. This indicates that higher input sizes do not enhance the precision of the bounding boxes, which can also be seen in figure \ref{fig:IoU} that depicts the average IoU for each input size. The value of mAP90 (at IoU=0.9) is close to zero, which shows that YOLOv3 has difficulties in getting the bounding boxes to be perfectly aligned with the object, as already noticed in the original YOLOv3 paper\cite{YOLOv3}.

On another hand, the increase in inference time for higher input sizes is due to the fact that a larger input size in the network generates a higher number of parameters, which raises the number of operations necessary to obtain the output.

Figure \ref{fig:TP-FP-FN} depicts the number of true positives (TP), false positives (FP), and false negatives (FN) for an IoU threshold of 0.5, and an input size of 320x320. We notice that the main issue with the ruku and sujud classes is the relatively high number of false negatives compared to the two other classes. This indicates that we should enrich the training dataset with a higher number of diverse images, especially for these two classes.

\section{CONCLUSIONS}

In this paper, we conducted a performance evaluation study for detecting Islamic prayer postures. The detection was performed by training the YOLOv3 network (with different configurations) on a dataset that was especially collected for the purpose. The performance of the different network configurations was assessed using several metrics (mAP, IoU, Inference time, TP, FP, and FN). The mAP at IoU=0.5 ranges from 78\% (for a network input size of 320x320) to 85\% (for an input size of 608x608), which is very encouraging for a training dataset of only 764 images.

This study is a first step that will be further developed in future works, by enlarging the training dataset, investigating other network architectures, conducting thorough hyperparameter optimization, and integrating the object detection model into a broader artificial intelligence assistive framework that aims to guide Muslim worshippers to evaluate the correctness of their postures during prayer.

\addtolength{\textheight}{-12cm}   





\section*{ACKNOWLEDGMENT}
This work is supported by the Robotics and Internet-of-Things Lab of Prince Sultan University. 

\FloatBarrier
\bibliographystyle{ieeetr}
\bibliography{ref}

\end{document}